\pdfoutput=1

\documentclass[11pt]{article}

\usepackage{amsmath}
\usepackage{amssymb}
\usepackage{mathtools}
\usepackage{amsthm}
\usepackage{enumitem}
\usepackage{algorithm}
\usepackage{algorithmic}
\usepackage{multicol}

\usepackage[preprint]{acl}

\usepackage{times}
\usepackage{latexsym}

\usepackage[T1]{fontenc}

\usepackage[utf8]{inputenc}

\usepackage{microtype}

\usepackage{inconsolata}

\usepackage{arydshln}
\usepackage{wrapfig}
\usepackage{multirow}
\usepackage{tabularx}
\usepackage{enumitem}
\usepackage{subcaption}
\usepackage{svg}
\usepackage{caption}
\usepackage{booktabs}
\usepackage{colortbl}
\usepackage{bbding}
\usepackage{float}
\usepackage{makecell}
\usepackage{tabularray} 
\usepackage{physics}

\usepackage{graphicx}
\newcommand{\beq}{\vspace{0mm}\begin{equation}}
\newcommand{\eeq}{\vspace{0mm}\end{equation}}
\newcommand{\beqs}{\vspace{0mm}\begin{eqnarray}}
\newcommand{\eeqs}{\vspace{0mm}\end{eqnarray}}
\newcommand{\barr}{\begin{array}}
\newcommand{\earr}{\end{array}}

\usepackage{booktabs}
\usepackage{array}
\newcolumntype{L}[1]{>{\raggedright\let\newline\\\arraybackslash\hspace{0pt}}m{#1}}
\newcolumntype{C}[1]{>{\centering\let\newline\\\arraybackslash\hspace{0pt}}m{#1}}
\newcolumntype{R}[1]{>{\raggedleft\let\newline\\\arraybackslash\hspace{0pt}}m{#1}}
\usepackage{geometry}
\usepackage{xcolor}
\usepackage[most]{tcolorbox}

\definecolor{mypromptgrayheader}{RGB}{50, 50, 50}      
\definecolor{mypromptbluebody}{RGB}{235, 243, 255}

\theoremstyle{plain}

\theoremstyle{definition}

\theoremstyle{remark}

\title{LegalDrill: Diagnosis-Driven Synthesis for Legal Reasoning in Small Language Models}

\author{
\textbf{Tianchun Li}\textsuperscript{1},
\textbf{Haochen Liu}\textsuperscript{2},
\textbf{Vishwa Pardeshi}\textsuperscript{2},
\textbf{Xingchen Wang}\textsuperscript{1},\\
\textbf{Tianci Liu}\textsuperscript{1},
\textbf{Huijun Zhao}\textsuperscript{2},
\textbf{Wei Fan}\textsuperscript{2},
\textbf{Jing Gao}\textsuperscript{1}\\[6pt]
\textsuperscript{1}Purdue University, West Lafayette, IN, USA\\
\textsuperscript{2}Fidelity Investments, Boston, MA, USA\\[6pt]
\texttt{\{li2657, wang2930, liu3351, jinggao\}@purdue.edu}\\
\texttt{\{haochen.liu, vishwa.pardeshi, huijun.zhao, wei.fan\}@fmr.com}
}

\begin{document}
\maketitle
\begin{abstract}
Small language models (SLMs) are promising for real-world deployment due to their efficiency and low operational cost. However, their limited capacity struggles with high-stakes legal reasoning tasks that require coherent statute interpretation and logically consistent deduction. Furthermore, training SLMs for such tasks demands high-quality, concise reasoning trajectories, which are prohibitively expensive to manually collect and difficult to curate via standard rejection sampling, lacking granularity beyond final verdicts. 
To address these challenges, we propose {LegalDrill}, a diagnosis-driven synthesis framework that extracts and iteratively refines reasoning trajectories from a capable teacher via fine-grained prompting, then a self-reflective verification is employed to adaptively select the most effective data for the SLM student. The resulting data empower SLM training through supervised fine-tuning and direct preference optimization. Extensive experiments on several legal benchmarks demonstrate that {LegalDrill} significantly bolsters the legal reasoning capabilities of representative SLMs while bypassing the need for scarce expert annotations, paving a scalable path toward practical legal reasoning systems.

\end{abstract}

\section{Introduction}
Large Language Models (LLMs) \cite{vaswani2017attention, radford2019language, brown2020language, comanici2025gemini} have demonstrated remarkable capabilities to understand, reason, and generate text \cite{gu2024survey, nam2024using, guo2025deepseek}, pushing new boundaries across a wide range of domains. As one of the most critical applications, there has been a growing trend to leverage LLMs for legal  domains in legal document retrieval \cite{siino2025exploring}, judgment prediction \cite{wu2023precedent}, and complex legal question answering \cite{guha2023legalbench}.

Despite their powerful capabilities, legal-domain frequently handles sensitive personal and confidential information. Routing such data to external APIs (e.g., GPT or Gemini) or cloud-based retrieval-augmented generation (RAG) pipelines introduces unacceptable privacy and security risks  \cite{siino2025exploring}. Therefore, it is strictly required that models be deployed within secure, local environments. The larger variants of open-sourced LLMs, such as Qwen3-32B \cite{yang2025qwen3technicalreport} or Llama3-70B \cite{grattafiori2024llama}, can be deployed locally but require expensive computational resources. Consequently, Small Language Models (SLMs)\footnote{In this work, we refer to SLMs as models with a parameter size of less than 3B.} have emerged as the pragmatic choice for secure, on-device deployment for legal applications \cite{lu2024small}.

However, the utility of SLMs remains limited in legal domains due to their weak reasoning ability \cite{fei2024lawbench, yu2025benchmarking}. Legal reasoning is not merely retrieving relevant clauses, but coherently interpreting statutes under case-specific (often ambiguous) contexts and carrying out logically valid deductions \cite{levi2022introduction, fan2025lexam, ioannou2025evaluating}. In practice, current SLMs often generate fluent, lawyer-like narratives, yet their reasoning is often fragile. They frequently make subtle errors—such as misreading statutory terms or making logical leaps—that undermine their final arguments. This gap is fundamentally tied to limited model size: constrained by their parameter scale, SLMs struggle to represent and execute the multi-step, dependency-heavy reasoning required for statute interpretation and logically consistent deduction.

Ideally, fine-tuning SLMs for high-stakes legal reasoning requires \emph{high-quality and concise} reasoning trajectories \cite{li2025small, wang2026decoupling}. Since manually collecting such legal reasoning data at scale is prohibitively expensive, a pragmatic alternative is to leverage stronger LLMs to generate reasoning traces via rejection sampling, e.g., selecting samples that match the final verdict or satisfy basic formatting constraints. However, this approach faces a fundamental behavioral and learnability mismatch between LLMs and SLMs \cite{ranaldi-freitas-2024-aligning,guo2025deepseek, yeo2025demystifying}. Reinforcement Learning (RL)-aligned LLMs typically rely on verbose, self-corrective deliberations—exhaustively exploring alternatives and revisiting premises—to ensure correctness \cite{jaech2024openai, guo2025deepseek}. In contrast, recent findings suggest that SLMs, constrained by their limited parameter scale, cannot effectively internalize or benefit from learning and imitating such ultra-long reasoning chains \cite{liu2025understanding, li2025small, yang2025understanding}. This discrepancy renders standard rejection sampling insufficient for the legal domain: it fails to curate legal reasoning paths that are concise enough for SLMs to learn.

To bridge this behavioral gap and enable SLMs to \textit{learn effectively}, we propose LegalDrill, an iterative framework designed to synthesize high-quality, concise reasoning trajectories tailored to the capacity of the student model. Instead of generating verbose, exploration-heavy chains that overwhelm smaller models from stronger models, we employ a diagnosis-driven mechanism. An audit Agent first scrutinizes the SLM's errors to pinpoint root causes, such as statutory misinterpretation or logical leaps. Guided by this diagnosis, a stronger model synthesizes a preference pair: a \textit{rejected} response mimicking the specific error pattern, and a \textit{chosen} response that rectifies it through a \textit{concise}, logically tight deduction. This process effectively converts the implicit knowledge of LLMs into compact, focused reasoning paths that are explicitly aligned with the SLM's learning behavior.

Furthermore, to ensure that the SLM learns efficiently, we introduce a self-reflective verification mechanism to filter out trivial samples. The preference pair generated by a stronger model might be objectively high-quality but subjectively trivial if the SLM is already capable of recognizing the correct reasoning. To address this mismatch, we leverage the SLM's own probability distribution to calculate a Difficulty Score that quantifies the model's confusion level, which is then used to filter the synthesized pairs. Specifically, this score identifies instances where the SLM incorrectly assigns higher confidence to the wrong reasoning (rejected response) over the corrected one (chosen response). By retaining these \textit{confused} samples, we curate a highly targeted training set that focuses exclusively on the model's actual blind spots. Finally, the SLM is optimized on these verified pairs via Supervised Fine-Tuning (SFT) and Direct Preference Optimization (DPO) \cite{rafailov2023direct}.

The contribution of our work is summarized as follows: First, we propose a diagnosis-driven synthesis framework that translates the implicit knowledge of strong LLMs into concise, error-correcting reasoning traces specifically tailored to the SLM's capacity. Second, we introduce a self-reflective verification mechanism that filters training data based on the student model’s intrinsic confusion, ensuring that the model focuses on its actual blind spots. Finally, our framework demonstrates effectiveness not only on open benchmarks but also in a \textit{real-world industrial setting} with experiments on proprietary datasets, comprising complex legal documents and financial contracts.

\section{Preliminaries}\label{sec:preliminary}
\subsection{Legal Task Formulation}
Legal reasoning requires analytical explanation and concept interpretation. For a given dataset $\mathcal{D}$, we denote the input as $x=(c,q)$, where $c$ is the legal context and $q$ is the query. Given $x$, the model generates an output sequence $y$ that contains the reasoning and, when applicable, the final answer. Depending on the task, $y$ may include an explicit verdict/prediction (e.g., judgment prediction) or consist purely of explanation (e.g., concept interpretation). Unless stated otherwise, we treat $y$ as a single sequence and omit the verdict $a$.

\subsection{Iterative Preference Optimization for Reasoning}
Direct Preference Optimization (DPO) \cite{rafailov2023direct} and its variants \cite{azar2024general, melnyk2024distributional, d2025anchored, jung2025binary}, instead of relying on an explicit reward model, directly uses pair-wise preference data to optimize the policy model with an equivalent optimization objective. Specifically, for a preference triple $(x, y^{+}, y^{-}) \sim \mathcal{D}$, DPO minimizes the following objective
\begin{align*}
\mathcal{L}_{\mathrm{DPO}}(\theta)
&= -\mathbb{E}_{(x, y^{+}, y^{-}) \sim \mathcal{D}}
\Big[\\
&\log \sigma\big(\beta(\log \tfrac{\pi_{\theta}(y^{+}\mid x)}{\pi_{\mathrm{ref}}(y^{+}\mid x)} - \log \tfrac{\pi_{\theta}(y^{-}\mid x)}{\pi_{\mathrm{ref}}(y^{-}\mid x)})\big)\Big],
\end{align*}
where $\pi_{\theta}$ is the policy model, $\pi_{\mathrm{ref}}$ is the fixed reference model, $\sigma(\cdot)$ is the sigmoid function, and $\beta$ is a parameter controlling the deviation from the reference model.  

Recent works \cite{pang2024iterative, deng2024flow, guo2024direct, tu2025enhancing, xu2025full} suggest that the reasoning abilities of language models could also be improved through an iterative online DPO. The training process will take $T$ rounds. At each iteration, the current model $\pi_{\theta_t}$ interacts with the environment (or a stronger teacher model) to generate a preference dataset $\mathcal{D}^t$. The model is then updated to  $\pi_{\theta_{t+1}}$ by minimizing the $\mathcal{L}_\mathrm{DPO}$ on $\mathcal{D}^t$.

\section{Methodology}
\begin{figure*}[htbp]
    \centering
    \includegraphics[width=0.9\linewidth, trim=0 6cm 0 6.5cm, clip]{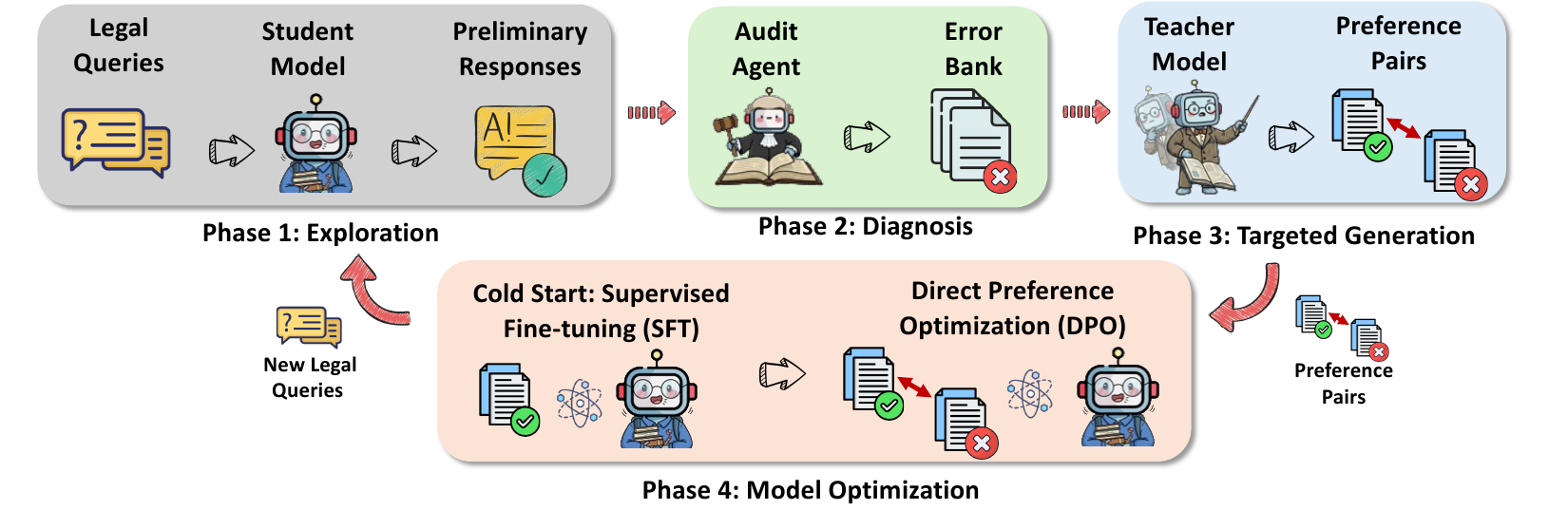}
    \vspace{-2mm}
    \caption{The overview of LegalDrill.}
    \vspace{-5mm}
    \label{fig:overview}
\end{figure*}

As illustrated in Fig.~\ref{fig:overview}, we propose an iterative framework designed to progressively refine the legal reasoning capabilities of SLMs. In the following sections, we refer to the stronger LLM as the \textbf{\textit{teacher model}} and the target SLM as the \textbf{\textit{student model}}. At each iteration $t$, we first synthesize reasoning trajectories that are concise and explicitly aligned with the student's behavior (Section~\ref{sec:generation}). We then employ a verification mechanism to filter these trajectories based on the student's current capabilities, ensuring only non-trivial samples are retained (Section~\ref{sec:verification}). Finally, in Section~\ref{sec:optimization}, we leverage this curated dataset to update the student model's parameters.

\subsection{Agent-Driven Instruction Refinement and Preference Data Generation}\label{sec:generation}
To bridge the gap between LLM capabilities and SLM constraints, we propose an iterative data synthesis framework. The core intuition is to diagnose the SLM's specific error patterns and synthesize concise, corrective reasoning traces that directly address these weaknesses. We structure this process into three stages: \textit{Exploration}, \textit{Diagnosis}, and \textit{Targeted Generation}.

\textbf{\textit{Exploration}}: 
Given a portion of the training data consisting of $N$ legal queries, the \emph{Exploration} stage prompts the current student model $\pi_{\theta_t}$ to generate a preliminary response $\hat{y}_i \sim \pi_{\theta_t}(\cdot \mid x_i)$ for each query $x_i$. 
To further encourage detailed reasoning traces, we utilize the Chain-of-Thought (CoT) system prompt, forcing the model to explain its internal logic and thereby making latent reasoning errors observable.

\textbf{\textit{Diagnosis}}: For \textit{Diagnosis}, we employ a specialized \textit{Audit Agent} ($\pi_{\text{audit}}$) to scrutinize each student response $\hat{y}_i$. The Agent identifies root causes of reasoning errors (e.g., misinterpretation of statutes or logical leaps) and abstracts them into generalized \textit{error simulation instructions}. To ensure that the diagnosis aligns with industry standards, we inject a taxonomy of common legal mistakes into the Agent's prompt. Formally, for each sample in the batch, we let the Agent to generate an instruction:
\begin{equation*}
    \mathcal{I}^{(i)} \sim \pi_{\text{audit}}(\cdot \mid \textit{Agent Prompt}(x_i, \hat{y}_i)).
\end{equation*}
Crucially, we constrain the Agent to generate \textit{context-agnostic} instructions (e.g., ``Ignore the time frame when calculating the limitation period''). This decoupling allows us to compile a reusable \textit{Error Instruction Bank}, denoted as $\Phi_{\text{err}} = \{\mathcal{I}^{(1)}, \dots, \mathcal{I}^{(N)}\}$, where instructions are no longer bound to their original source contexts.


\textbf{\textit{Targeted Generation}}: Finally, we leverage the \textit{Error Instruction Bank} $\Phi_{\text{err}}$ to synthesize preference data at scale. The key motivation is two-fold. First, because each instruction $\mathcal{I}$ is \textit{context-agnostic}, the same reasoning failure can be reproduced under many different legal contexts. This acts as a regularizer against shortcut learning: the contrast between the chosen and rejected responses is driven by reasoning rigor rather than superficial cues (e.g., response length or lexical patterns). Second, decoupling \textit{error types} from \textit{contexts} enables combinatorial expansion: we can generate arbitrarily many training pairs by recombining contexts with diverse error instructions, instead of being constrained by the size of the original dataset.

Concretely, for each training sample, we sample $K$ error instructions $\{\mathcal{I}_k\}_{k=1}^{K}$ from $\Phi_{\text{err}}$, where $K$ is a hyperparameter controlling the expansion ratio. For each instruction $\mathcal{I}_k$, a stronger \textit{teacher model} ($\pi_{\text{teach}}$) synthesizes one preference pair by a two-step procedure.

For each instruction $\mathcal{I}_k$, the teacher model first generates a \textit{rejected response} by intentionally following the specified erroneous logic,
\begin{equation*}
    y_-^{(k)} \sim \pi_{\text{teach}}(\cdot \mid x, \mathcal{I}_k).
\end{equation*}
To generate a high-quality \textit{chosen} response in a more targeted way, 
we further provide the teacher model with this paired \emph{rejected} response. 
The generation process is given by

\begin{equation*}
    y_+^{(k)} \sim \pi_{\text{teach}}(\cdot \mid x, \mathcal{I}_k, y_-^{(k)}).
\end{equation*}

Overall, each original sample $x$ yields $K$ preference pairs, leading to a synthesized dataset of size $K\cdot |\mathcal{D}|$:
\begin{equation*}
    \mathcal{D}^t_{\text{syn}} = \{(x, y_+^{(k)}, y_-^{(k)}) \mid x \in \mathcal{D},\; k = 1,\dots,K\}.
\end{equation*}

\subsection{Self-Reflective Quality Verification}\label{sec:verification}
The synthetic pairs $\mathcal{D}_{\text{syn}}^t$ generated in the previous stage may contain samples that are trivial if the student model $\pi_{\theta_t}$ can already determine the correct reasoning. To focus optimization on the model's actual blind spots, we introduce a \textit{Self-Reflective Verification} mechanism. This process filters the dataset to retain only those pairs where the student genuinely struggles to differentiate the correct reasoning trajectory from the error.

We propose the \textit{Difficulty Score} (DS) to quantify the alignment-or-conflict between the student model's belief and the teacher-synthesized preferences. Importantly, we do not estimate this belief via the likelihood $\pi_{\theta_t}(y\mid x)$ over the entire sequence, which could be sensitive to response length and surface form. Instead, we leverage the student's instruction-following capability to perform a forced-choice prediction task.

Concretely, given a legal context $c$, query $q$, and a candidate response $y$, we construct a structured verification prompt $\mathcal{P}_{\text{ver}}(c, q, y)$ that mirrors the formulation of the legal task described in Sec.\ \ref{sec:preliminary}. We strictly constrain the output words to a binary set $\mathcal{V} = \{\texttt{correct}, \texttt{incorrect}\}$ in the prompt $\mathcal{P}_{\text{ver}}(c, q, y)$. Rather than relying on raw probabilities, we normalize the prediction confidence over these two words to isolate the model's prediction. We define the normalized correctness score as:
\begin{equation*}
    \scalebox{0.85}{
        $\displaystyle
        s_{\theta_t}(y \mid x) = \frac{\pi_{\theta_t}(\texttt{correct} \mid \mathcal{P}_{\text{ver}})}{\pi_{\theta_t}(\texttt{correct} \mid \mathcal{P}_{\text{ver}}) + \pi_{\theta_t}(\texttt{incorrect} \mid \mathcal{P}_{\text{ver}})}
        $
    }
\end{equation*}

This normalization ensures that even if the absolute probability of generating the specific token ``correct'' fluctuates during the iterative training, the \textit{relative} preference remains a valid signal of the model's internal belief. For each preference pair $(x, y_+^{(k)}, y_-^{(k)})$, we simply compute the Difficulty Score as the margin between the student's endorsement of the error and the correction:
\begin{equation*}
    \text{DS}(x, y_+^{(k)}, y_-^{(k)}) = s_{\theta_t}(y_-^{(k)} \mid x) - s_{\theta_t}(y_+^{(k)} \mid x).
\end{equation*}

The DS measures how strongly the student is \emph{misled} by the rejected reasoning relative to the chosen one. If $\text{DS}<0$, the student identifies the correct reasoning $y_+^{(k)}$, making the pair relatively easy. If $\text{DS}>0$, the student explicitly assigns higher confidence to the flawed reasoning $y_-^{(k)}$, revealing a genuine blind spot (or ``confusion''). Based on this metric, we construct the final training set $\mathcal{D}_{\text{train}}^t$ by strictly filtering for high-value samples. We apply a thresholding strategy where only pairs satisfying $\text{DS} > \tau$ are retained. This filtering focuses the optimization budget exclusively on logical fallacies where the student model is most vulnerable.

\subsection{Optimization Objectives: SFT and DPO}\label{sec:optimization}
After targeted generation and self-reflective filtering, we obtain a training set of preference triples $(x, y_+^{(k)}, y_-^{(k)}) \in \mathcal{D}_{\text{train}}^t$. We optimize the student model in two stages. First, when $t=0$ we perform supervised fine-tuning on the teacher-preferred responses to provide a stable cold start for preference optimization. Then, we apply DPO on the filtered preference pairs.

\paragraph{Supervised fine-tuning (cold start).}
When $t=0$, we initialize the student by maximizing the likelihood of the chosen responses:
\begin{equation*}
    \mathcal{L}_{\text{SFT}}(\theta_0) = -\mathbb{E}_{(x, y_+) \sim \mathcal{D}_{\text{train}}^0}\big[\log {\pi_{\theta_0}}(y_+ \mid x)\big].
\end{equation*}
This warm-up stage anchors the model on high-quality reasoning traces before applying the pairwise preference objective.

\begin{figure*}[htbp]
    \centering
    \includegraphics[width=1.0\linewidth, trim=0 3.5cm 0 3cm, clip]{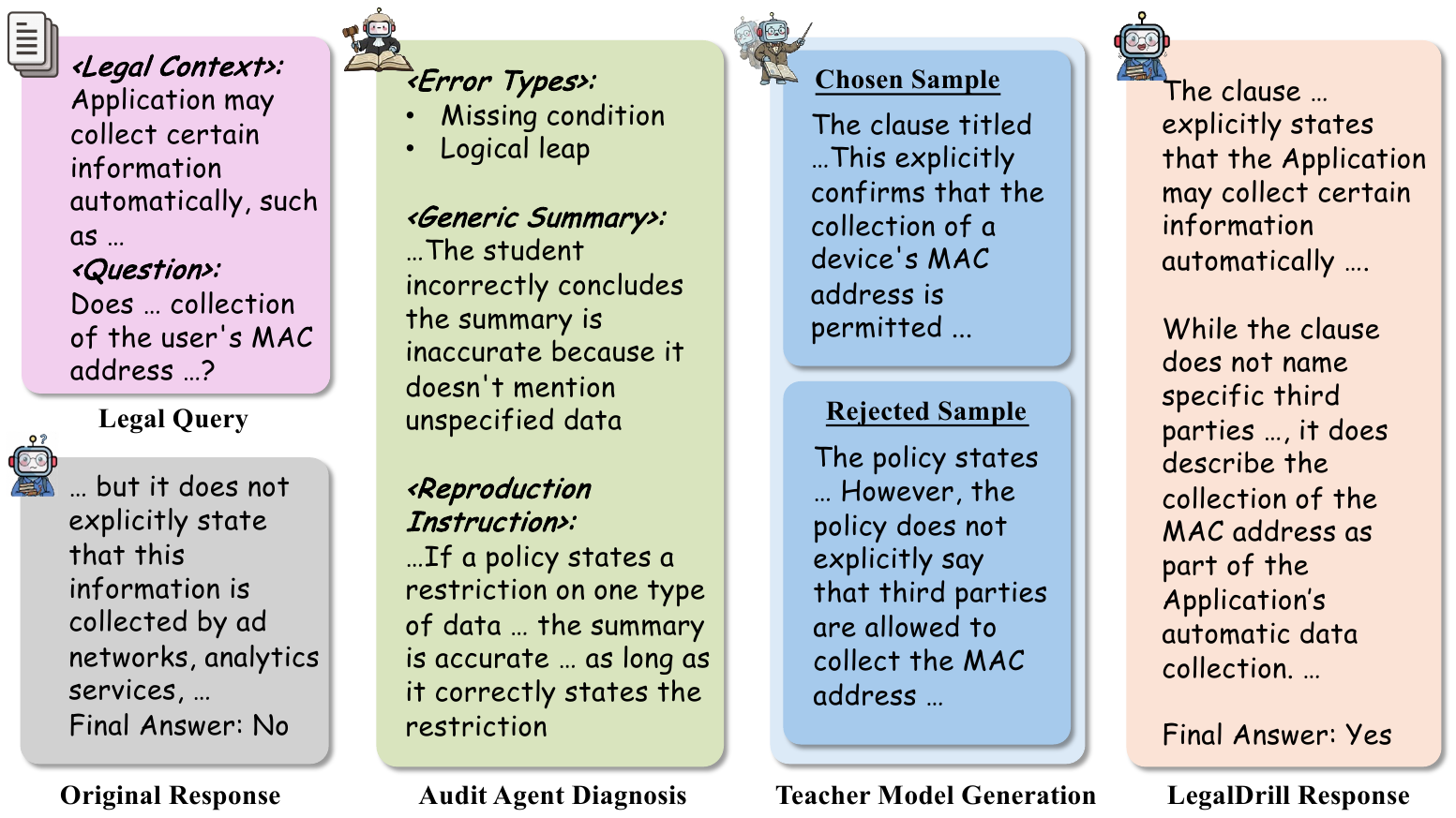}
    \caption{Illustrative examples of legal reasoning refinement of LegalDrill. After the optimization, the student model is asked to generate a response under the same legal query, which is marked as LegalDrill response}
    \vspace{-5mm}
    \label{fig:casestudy}
\end{figure*}

\paragraph{Direct preference optimization.}
Following the SFT warm-up (or for subsequent iterations $t \geq 0$), we refine the model by minimizing the preference objective $\mathcal{L}_{\text{DPO}}$. Specifically, we update the policy from $\pi_{\theta_t}$ to $\pi_{\theta_{t+1}}$ using the filtered dataset $\mathcal{D}_{\text{train}}^t$. During the iterative process, we set the reference model $\pi_{\text{ref}}$ to be the current policy $\pi_{\theta_t}$. 
\section{Experiment}
\subsection{Experimental Setup. } 
\paragraph{Backbone models, teacher models, and agent models}We train our student models with two SLM backbones: Qwen3-0.6B and Qwen3-1.7B. For both the teacher model and the audit agent, we consistently use the same stronger model, choosing between the open‑sourced Qwen3‑30B‑A3B‑Instruct and the closed‑sourced GPT‑4o.


\paragraph{Baselines.} We evaluate performance using zero-shot CoT prompting across two categories: (1) \textbf{General LLMs}, including the teacher models (GPT-4o, Qwen3-30B-Instruct) and the student base models (Qwen3-0.6B, Qwen3-1.7B); and (2) \textbf{Legal-Specific LLMs}, namely Law-LLM-13B \cite{cheng2023adapting}, DeepSeek ESFT-16B \cite{wang2024let}, and DiscLaw-13B \cite{yue2023disc}.

\paragraph{Datasets} We evaluate our method on six datasets: four public legal‑reasoning benchmarks from LegalBench \cite{guha2023legalbench}: \textit{Consumer QA}, \textit{Contracts QA}, \textit{Sara Entailment}, and \textit{Privacy Policy Entailment}, as well as two proprietary datasets from real‑world financial‑industry legal document review scenarios, namely \textit{Real‑World power of attorney (POA)} and \textit{Real‑World Trust}. All datasets consist of document‑grounded, binary yes/no question–answer pairs. Detailed descriptions of both the public benchmarks and the proprietary datasets are provided in appendix \ref{app:datasets}.

\paragraph{Metrics.} We report accuracy and F1 to evaluate each model’s judgment performance on the binary QA tasks. In addition, we introduce a “judge accuracy” metric to assess the quality of the generated reasoning, using an LLM as a Judge. Implementation details are provided in appendix \ref{app:judge_acc} and \ref{app:prompts}.


\begin{table*}[t]
  \centering
  \small
  \setlength{\tabcolsep}{3.5pt}
  \renewcommand{\arraystretch}{1.15}

  \caption{Overall performances on public datasets.}

  \begin{tabular}{l ccc ccc ccc ccc ccc}
    \toprule
      & \multicolumn{3}{c}{\textbf{Cos. QA}}
      & \multicolumn{3}{c}{\textbf{Con. QA}}
      & \multicolumn{3}{c}{\textbf{Sara Ent.}}
      & \multicolumn{3}{c}{\textbf{Priv. Ent.}}
      \\
      & \textit{Acc} & \textit{F1} & \textit{Judge}
      & \textit{Acc} & \textit{F1} & \textit{Judge}
      & \textit{Acc} & \textit{F1} & \textit{Judge}
      & \textit{Acc} & \textit{F1} & \textit{Judge}\\
    \midrule

    \multicolumn{16}{l}{\textbf{Law Model} Zero Shot}  \\
    DeepSeek ESFT-16B  &0.81  &0.80  &0.74  &0.88  &0.87  &0.79  &0.46  &0.44  &0.27  &0.75  &0.64  &0.73  &  &  &  \\
     Law-LLM-13B     &0.49  &0.38  &0.36  &0.88  &0.56  &0.77  &0.51  &0.38  &0.18  &0.64  &0.38  &0.29  &  &  &  \\
    DiscLaw-13B    &0.24  &0.18  &0.11  &0.42  &0.33  &0.38  &0.10  &0.07  &0.05  &0.01  &0.01  &0.01  &  &  &  \\
    \midrule

    \multicolumn{16}{l}{\textbf{General SLMs and LLMs} Zero Shot} \\
	
	Qwen3-0.6B &0.69 &0.34 &0.66  &0.83 &0.83 &0.75  &0.59 &0.27 &0.18  &0.30 &0.29 &0.24  & & & \\
	Qwen3-1.7B &0.79 &0.58 &0.70  &0.87 &0.85 &0.81  &0.66 &0.38 &0.37  &0.47 &0.45 &0.30  & & & \\
     Qwen3-30B-A3B-Instruct &0.98 &0.97 &0.92  &0.96 &0.95 &0.93  &0.86 &0.43 &0.51  &0.83 &0.75 &0.65  & & & \\
    GPT-4o  & 0.98 & 0.98 & 0.81 & 0.92 & 0.92 & 0.92 & 0.83 & 0.81 & 0.62 & 0.67 & 0.30 & 0.60 &  &  &  \\
    \midrule

    \multicolumn{16}{l}{\textbf{Ours: Distill from  Qwen3-30B-A3B-Instruct}} \\
    \rowcolor{gray!10}LegalDrill-0.6B  &0.84 &0.83 &0.77  &0.91 &0.88 &0.85  &0.74 &0.45 &0.44  &0.81 &0.80  &0.58  &  &  &  \\
    \rowcolor{gray!10}LegalDrill-1.7B  &0.96 &0.96  &0.89  &0.93 &0.92 &0.83  &0.73 &0.39  &0.42  &0.85  &0.80  &0.59  &  &  &  \\
    \midrule

    \multicolumn{16}{l}{\textbf{Ours: Distill from GPT-4o}} \\
    \rowcolor{gray!10}LegalDrill-0.6B      &0.86  &0.84  &0.75  &0.95  &0.84   &0.91  &0.75  &0.42  &0.41  &0.59  &0.32  &0.52  &  &   &  \\
    \rowcolor{gray!11} LegalDrill-1.7B  &0.94  &0.94  &0.88  &0.97  &0.94  &0.94  &0.75  &0.46 &0.43 &0.60  &0.31  &0.52  &  &  &\\
    \bottomrule
  \end{tabular}

  \label{tab:overall_performance}
\end{table*}

\begin{table}[h!]
  \centering
  \small
  \setlength{\tabcolsep}{3.5pt}
  \renewcommand{\arraystretch}{1.15}

  \caption{Performances on Real-World datasets.}
  \vspace{-2mm}

  \begin{tabular}{l cc cc c c c c c c}
    \toprule
      & \multicolumn{3}{c}{\textbf{Real-World POA}}
      & \multicolumn{3}{c}{\textbf{Real-World Trust}}\\
      & \textit{Acc} & \textit{F1} & \textit{Judge}
      & \textit{Acc} & \textit{F1} & \textit{Judge}\\
    \midrule

    \multicolumn{4}{l}{\textbf{General SLMs and LLMs}} \\
	Qwen3-0.6B &0.76 & 0.51 & 0.27 & 0.74 & 0.49 & 0.44\\
	Qwen3-1.7B &0.78 & 0.53 & 0.50 & 0.79 & 0.54 & 0.51 \\
    GPT-4o  & 0.91 & 0.90 & 0.67 & 0.89 & 0.60 & 0.73 \\
    \midrule

    \multicolumn{4}{l}{\textbf{Ours: Distill from GPT-4o}} \\
    \rowcolor{gray!10}LegalDrill 0.6B & 0.87 & 0.58 & 0.72 & 0.86 & 0.58 & 0.69  \\
    \rowcolor{gray!10}LegalDrill 1.7B & 0.92 & 0.91 & 0.73 & 0.90  & 0.60 & 0.70  \\
    \bottomrule
  \end{tabular}
  \vspace{-5mm}

  \label{tab:realworld_performance}
\end{table}





\subsection{Quantitative Results}
Table~\ref{tab:overall_performance} reports the main results on four public LegalBench benchmarks as well as two in-domain document QA benchmarks from Real-World. We compare against both (i) zero-shot prompting of the teacher LLMs and (ii) undistilled student backbones. Overall, {LegalDrill} consistently improves both accuracy and F1 over the Qwen3‑0.6B/1.7B baselines across most datasets, indicating that error‑driven preference data effectively transfers the teachers’ decision boundary and reasoning behaviors to substantially smaller models.

On LegalBench, LegalDrill notably boosts the student models on contract QA tasks (Consumer QA, Contracts QA) and improves entailment tasks in both accuracy and F1, suggesting more reliable binary decisions under class imbalance. Importantly, distillation narrows the teacher–student gap: with Qwen3‑30B‑A3B‑Instruct as teacher, the 1.7B student reaches near‑teacher performance on multiple benchmarks. On Real‑World datasets, distilling from GPT‑4o yields student performance on par with the teacher, enabling cost‑effective deployment for automated compliance workflows in the industry. Beyond final yes/no correctness, the "judge accuracy" further improves over the Qwen3 backbones on most datasets, indicating fewer judge‑detected errors and more reliable legal rationales.

We observe a consistent scaling trend: across zero‑shot and distilled settings, the 1.7B backbone outperforms the 0.6B backbone, with distillation amplifying this advantage. In several cases (e.g., contract QA), the 1.7B LegalDrill model becomes comparable to the Qwen3‑30B‑A3B‑Instruct teacher while being substantially lighter. This highligts that our method yields even greater gains on small models with moderately larger size, enabling them to approach the performance of large models.

\subsection{Qualitative Results}

We further present qualitative results to demonstrate how LegalDrill refines the legal reasoning and corrects the student's weak spots. After optimization, the student model is asked to generate a response for the same legal query. 
Fig.~\ref{fig:casestudy} illustrates the effect of LegalDrill on a randomly selected example. 
The full, detailed responses are included in App.~\ref{app:case_study}. 
Given the legal query, the student responds with reasoning and a conclusion that appear logical on the surface. 
However, the response inherently contains an error because the final yes/no conclusion is incorrect. 
Next, the Audit Agent analyzes potential errors in the reasoning and categorizes them into two error types. 
Furthermore, the Audit Agent generates instructions on how to reproduce the error under \textit{any} legal context by deliberately avoiding context-specific details, only mentioning terms like ``policy'' or ``one type of data''. 
Then, based on the analysis, the teacher model is instructed to generate a targeted pair, which reproduces and corrects the error. 
In the example, the teacher model pinpoints the key step in the reasoning and flips the logic between the chosen and rejected responses. 
After optimization, to demonstrate the effect of LegalDrill, the student model is asked to generate a response for the same legal context. 
The response from the student model demonstrates that the SLM clearly resolves the previous error and does not simply repeat the chosen response generated by the teacher model.

\subsection{Ablation Study}


We further validate the effectiveness of the DPO technique via an ablation study. As shown in Figure~\ref{fig:ablation}, across all settings the model trained with DPO consistently achieves higher accuracy than the SFT-only counterpart, which indicates that leveraging both chosen and rejected responses provides strong contrastive signals. Notably, the DPO almost universally improves the student model's reasoning robustness.

We refer more experiment results in App. \ref{sec:more_experiments} with the effects of number of iterations, $K$, etc.

\begin{figure}[htbp]
    \centering
    \includegraphics[width=\linewidth]{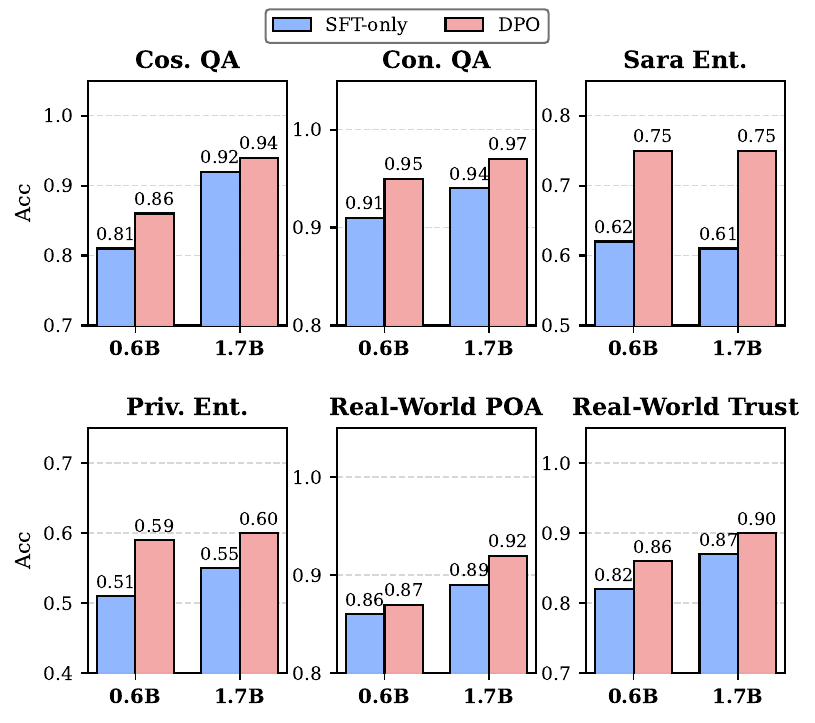}
    \vspace{-8mm}
    \caption{The ablation study on DPO.}
    \vspace{-5mm}
    \label{fig:ablation}
\end{figure}
\section{Related Work}

\textbf{LLMs for Legal Reasoning} Recent works \cite{cui2023chatlaw, wang2024let, zhou2024lawgpt, dai2025legal, shi2025legalreasoner, cai2025unilaw, li2025legal} have explored adapting LLMs to the legal domain. Most recent works adopt supervised fine-tuning and reinforcement learning strategies. Unlike rewards in general domains \cite{zheng2025survey, gunjal2025rubrics, liu2025openrubrics, xu2026alternating}, the legal domain requires domain-specific rewards to enable models to accurately handle legal queries. To name a few, \citeauthor{shi2025legalreasoner} incorporate progressiveness and potential, which describe the step-wise reward for intermediate legal reasoning steps. \citeauthor{cai2025unilaw} design a legal validity reward to encourage legal accuracy. \citeauthor{dai2025legal} formulate the legal reward from an information theory perspective. However, even with Parameter-Efficient Fine-Tuning (PEFT) methods \cite{hu2022lora, zhang2023adalora, wu2024reft, liu2025roserag} and quantization or pruning strategies, LLMs still consume a considerable amount of resources to train and deploy.

\textbf{Knowledge Distillation and Reasoning Distillation} Knowledge distillation \cite{hinton2015distilling, phuong2019towards, gou2021knowledge, tan2023gkd} focuses on transferring knowledge from a larger teacher model to a smaller student model by aligning their output logits or intermediate representation features. With the advancement of language models, the focus of distillation has shifted towards distilling complex, multi-step reasoning capabilities from LLMs into SLMs \cite{shridhar2023distilling, kang2023knowledge,feng2024teaching, zhao2025can, yang2025qwen3technicalreport, zhang2025lightthinker, kim2025guiding}. However, standard methods struggle with a fundamental behavioral mismatch: constrained by their parameter scale, SLMs cannot effectively internalize the verbose, self-corrective reasoning chains typical of strong LLMs. While recent studies explore trajectory refinement, methods such as reasoning compression \cite{zhao2025can, zhang2025lightthinker} primarily prune long CoT traces to bridge the reasoning gap between SLMs and LLMs. In contrast, LegalDrill does not merely compress reasoning chains; it generates trajectories specifically targeting the SLM's unique blind spots. Furthermore, unlike frameworks such as SMART \cite{kim2025guiding} that rely on external LLMs during inference, LegalDrill uses curated trajectories during training to resolve the model's logical blind spots, enabling the SLM to reason correctly without external inference-time intervention.

\section{Conclusion}
While LLMs demonstrate strength in legal reasoning, real-world legal deployment often requires privacy-preserving and cost-efficient solutions. 
SLMs are promising for real-world deployment to resource constrained devices due to their efficiency and low operational cost, therefore providing privacy guaranties for the legal domain. We proposed LegalDrill, a diagnosis-driven synthesis framework that 
translates the implicit knowledge of strong LLMs into concise, corrective reasoning traces explicitly tailored to the student model's capacity. Experiments on LegalBench and real-world datasets show consistent gains over baseline SLMs.

\section*{Limitations}
While LegalDrill effectively enhances SLMs, the framework employs the DPO algorithm. Inside DPO, there are some hyperparameters to tune, e.g., learning rate, weight decay, epochs, etc. This will require some additional tuning. However, we find that typically, $1-3$ epochs are enough with a learning rate of $1 \times 10^{-4}$. We include the range of each parameter in the App.

\section*{Acknowledgment}
This work is supported in part by the US National Science Foundation under grant NSF IIS-2141037. Any opinions, findings, and conclusions or recommendations expressed in this material are those of the author(s) and do not necessarily reflect the views of the National Science Foundation.

\bibliography{writing/reference}


\appendix
\onecolumn
\appendix
\section{More Experiment and Experimental details}\label{sec:more_experiments}
\subsection{Judge Accuracy}
\label{app:judge_acc}
Judge accuracy is computed by employing another LLM as a judge: given each model's generated legal reasoning, the judge checks whether the reasoning contains any potential error; if so, the entire reasoning is marked as incorrect. We then report the resulting accuracy as a proxy for reasoning quality. Specifically, we use the Qwen3-8B as the judge.

\subsection{The impact of number of iterations}
We study how the number of self-improvement iterations affects downstream performance. In general, we find that only a small number of iterations is needed: most of the gains are achieved within the first one to two iterations, after which additional iterations yield only marginal improvements. To validate this trend, we run our framework for four iterations using Qwen3-30B-A3B-Instruct as the teacher model on two open benchmarks.

\begin{figure}[htbp]
    \centering
    \includegraphics[width=\linewidth]{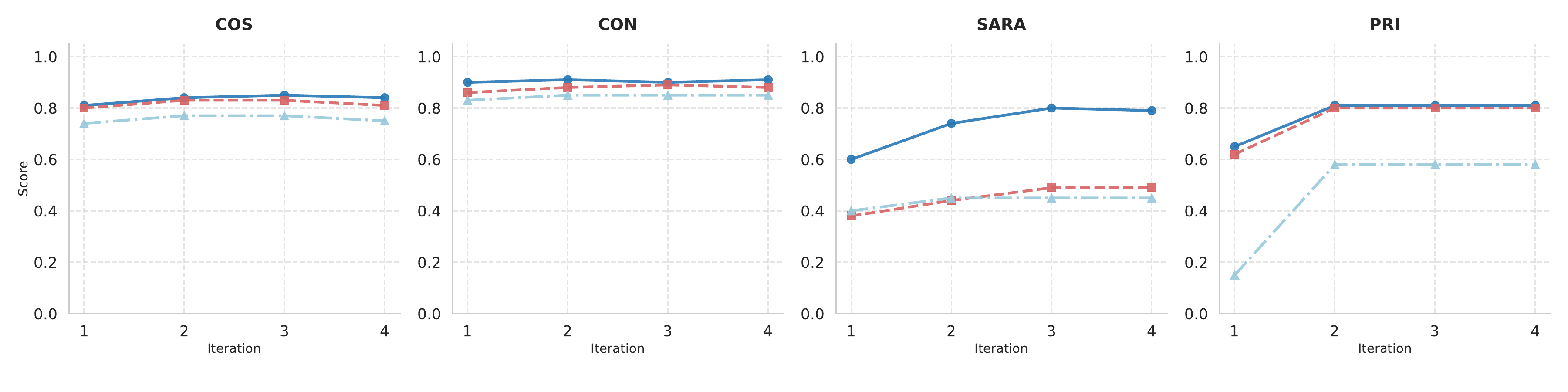}
    \vspace{-8mm}
    \caption{The ablation study on the number of iterations.}
    \vspace{-5mm}
    \label{fig:ablation}
\end{figure}

From the above results, we can see that the largest performance gain, typically the Judge score, happens around the second iteration. This actually confirms that the reasoning ability can be greatly improved after two iterations but will not keep improving. 
\subsection{The impact over $K$ preference pairs}
We investigate how the number of preference pairs $K$ used to construct the final training set affects performance. Overall, we observe that the distilled training set does not need to be large to be effective: in practice, a final training dataset of roughly 1{,}000--2{,}000 samples typically achieves the best results, while further increasing the data size brings only marginal gains. As a consequence, the appropriate $K$ depends on the size of the original legal dataset (and thus how many preference pairs can be harvested). Using Consumer QA (\textit{Cos. QA}) as an example, we sweep $K \in \{2,4,6,8,10,12,16\}$; when $K=16$, the resulting training set contains $\sim$3{,}000+ samples, and performance largely saturates within this range.
\begin{figure}[htbp]
    \centering    \includegraphics[width=0.6\linewidth]{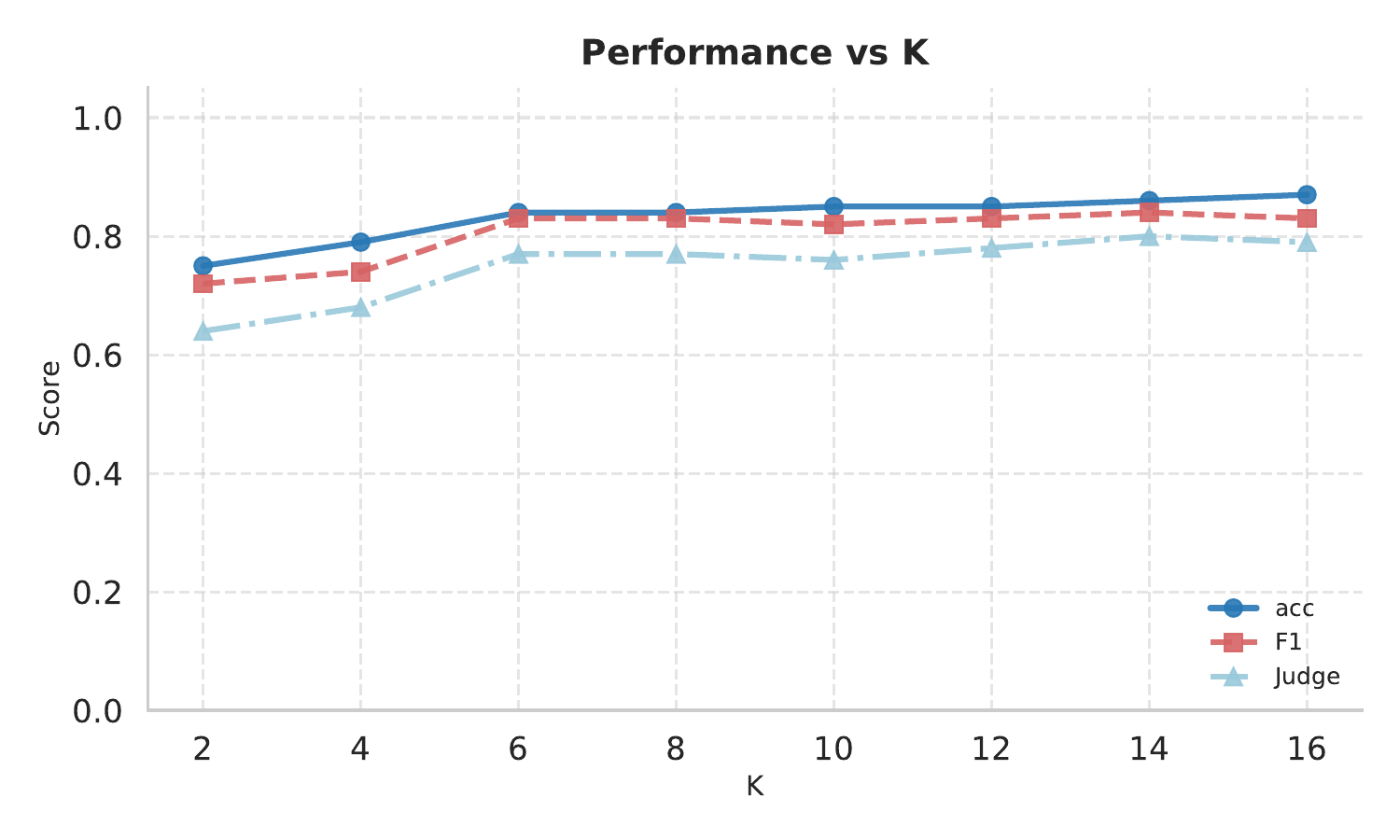}
    \vspace{-4mm}
    \caption{The ablation study on the number of K.}
    \vspace{-5mm}
    \label{fig:ablation}
\end{figure}

\subsection{LegalDrill Generation Demonstration}
\label{app:case_study}
\begin{tcolorbox}[
    enhanced,                   
    title={Response Generated by Qwen3 1.7B},
    colframe=mypromptgrayheader,         
    colback=mypromptbluebody,             
    coltitle=white,             
    fonttitle=\bfseries\large,  
    boxrule=0.5mm,              
    drop shadow,                
    sharp corners=downhill,   
    arc=3mm                 
]

Step 1: Identify the relevant clause in the contract. The contract mentions the ``Automatically Collected Information'' section, which states: ``In addition, the Application may collect certain information automatically, such as: the type of mobile device you use, mobile device identifiers, such as the resettable advertising identifier assigned to your device, the IP address of your mobile device, the location of your device, your mobile operating system, Wi-Fi mac address in your area, the type of mobile Internet browsers you use, and information about the way you use the Application. See `Automatic Data Collection and Advertising' section for examples.'' This section clearly outlines the types of information that may be collected automatically, including the Wi-Fi mac address in your area. Step 2: Determine if the MAC address is collected by ad networks, analytics services, or other third parties. The contract explicitly mentions that the Wi-Fi mac address in your area is included in the list of automatically collected information. It does not specify that this information is collected by ad networks, analytics services, or other third parties. While the contract does not explicitly state that the MAC address is collected by third parties, it does not deny that it may be collected by such entities. The clause is neutral in this regard and does not make a claim about the source of the data. Step 3: Evaluate whether the description accurately summarizes the clause. The question asks whether the policy accurately describes the clause. The clause does mention that the Wi-Fi mac address in your area is collected, but it does not explicitly state that this information is collected by ad networks, analytics services, or other third parties. Therefore, the description does not fully capture the intent of the clause, as it lacks a definitive statement about the source of the information. Final Answer: No

\end{tcolorbox}

The above is an example of the SLM's response. Note that while the step-by-step reasoning appears logical on the surface, it inherently contains an error because the final yes/no conclusion is wrong. In the next step, we will feed this response to the Audit Agent to conduct the error analysis.

\begin{tcolorbox}[
    enhanced,
    title={Audit Agent Instruction to Teacher Model},
    colframe=mypromptgrayheader,
    colback=mypromptbluebody,
    coltitle=white,
    fonttitle=\bfseries\large,
    boxrule=0.5mm,
    drop shadow,
    sharp corners=downhill,
    arc=3mm
]

\textbf{Status:} \\
\textbf{\texttt{INCORRECT\_ANSWER}}
\vspace{0.2cm}
\textbf{Error Types:}
\begin{itemize}[leftmargin=*, nosep]
    \item Missing condition
    \item Logical leap
\end{itemize}

\vspace{0.2cm}
\textbf{Generic Summary:} \\
The student correctly identifies that the clause permits collection of unspecified location data but fails to recognize that the policy description explicitly excludes only precise location data, thereby implying it may still collect unspecified data. The student incorrectly concludes the summary is inaccurate because it doesn't mention unspecified data, when in fact the policy's statement is accurate as written---because it only claims a restriction on precise data, which matches the clause.

\vspace{0.2cm}
\textbf{Reproduction Instruction:} \\
\textit{When evaluating a policy summary, ensure that the summary's claim does not contradict the contract's explicit language. If a policy states a restriction on one type of data (e.g., precise location), and the contract allows another (e.g., unspecified location), the summary is accurate even if it does not mention the permitted data, as long as it correctly states the restriction.}

\end{tcolorbox}

The above is an example of the Audit Agent's response. Note that only the Reproduction Instruction will later be used to generate a chosen and rejected sample. Notably, by properly instructing the agent through prompting, the agent is able to transform the error analysis into a generalizable reproduction instruction. This crucial step allows us to, firstly construct an error bank, and, secondly, to reproduce the SLM's behavior easily.

Then, we demonstrate the generated chosen and rejected sample from the Teacher model under a completely different legal context from the error bank.

\begin{tcolorbox}[
    enhanced,
    title={Chosen Sample Generated by Teacher},
    colframe=mypromptgrayheader,
    colback=mypromptbluebody,
    coltitle=white,
    fonttitle=\bfseries\large,
    boxrule=0.5mm,
    drop shadow,
    sharp corners=downhill,
    arc=3mm
]

The clause titled ``Automatically Collected Information'' states that the Application may collect certain information automatically, including ``Wi-Fi mac address in your area.'' This explicitly confirms that the collection of a device's MAC address is permitted as part of the automatic data collection process. The clause further clarifies that such data may be collected by third parties, including ad networks and analytics services, through the ``Automatic Data Collection and Advertising'' section, which is referenced in the text.

While the policy does not specify the exact context or method of collection by third parties, it does not restrict or prohibit the collection of MAC addresses by these entities. The absence of any explicit prohibition, combined with the clear mention of Wi-Fi MAC address as an automatically collected data point, supports the conclusion that such collection is permitted. The policy's description that ``the user's MAC address is collected by ad networks, analytics services, or other third parties'' aligns directly with this provision, as it reflects the actual scope of collection described in the clause.

There is no logical inconsistency or factual misrepresentation in the description, and no requirement to mention additional data types (such as unspecified location data) for the summary to be accurate. The clause only restricts precise location data, which is not relevant to the claim about MAC address collection. Therefore, the policy description is accurate and consistent with the terms of the clause.

\vspace{0.3cm}
\textbf{Final Answer: Yes}

\end{tcolorbox}

\begin{tcolorbox}[
    enhanced,
    title={Rejected Sample Generated by Teacher},
    colframe=mypromptgrayheader,
    colback=mypromptbluebody,
    coltitle=white,
    fonttitle=\bfseries\large,
    boxrule=0.5mm,
    drop shadow,
    sharp corners=downhill,
    arc=3mm
]

The policy states that the Application collects certain information automatically, including the ``Wi-Fi MAC address in your area.'' This means that third parties like ad networks and analytics services can access the MAC address when you are connected to Wi-Fi. However, the policy does not explicitly say that these third parties are allowed to collect the MAC address---only that the Application collects it.

Since the clause does not clearly state that third parties are permitted to access or use this data, it's a logical leap to assume that the policy allows such collection by external entities. Moreover, the policy only mentions that the Application collects the MAC address, not that it shares it with ad networks or analytics services. Therefore, the summary overreaches by implying that third parties are involved in collecting the MAC address, which is not confirmed by the clause. Because the policy does not specify that third parties collect the MAC address, the description is misleading and inaccurate.

\vspace{0.3cm}
\textbf{Final Answer: No}
\end{tcolorbox}

From above we can see that, without any legal expertise and any prior knowledge, it would be hard to clearly see which one is the chosen response. This highlights the complexity of legal reasoning. The rejected sample constructs a cohesive and syntactically fluent argument that appears entirely plausible on the surface. By accurately citing specific clauses (e.g., "Wi-Fi MAC address") while subtly distorting the interpretative scope regarding third-party permissions, it creates a deceptive facade of validity. The error here is not a simple factual hallucination, but a sophisticated "logical leap"—a failure to properly deduce implied permissions from the text's structure. Unlike obvious errors, this type of flawed reasoning adheres to the form of legal analysis while violating its substance. Consequently, distinguishing the correct interpretation requires not just textual alignment, but a rigorous verification of logical consistency between the claim and the contract terms.

The example below presents a response generated by LegalDrill 1.7B, which was distilled from the teacher model Qwen3-30B-A3B-Instruct. Despite its compact parameter size, the model successfully replicates the rigorous reasoning structure of the teacher, accurately identifying the trigger (material breach) and procedural requirements (cure period) within the clause. It moves beyond simple extraction to synthesize a coherent legal argument regarding the cessation of obligations. This demonstrates that our distillation framework effectively transfers not just linguistic fluency, but the underlying logic required for contract analysis.

\begin{tcolorbox}[
    enhanced,                   
    title={Response Generated by LegalDrill 1.7B (distilled from Qwen3-30B-A3B-Instruct)},
    colframe=mypromptgrayheader,         
    colback=mypromptbluebody,             
    coltitle=white,             
    fonttitle=\bfseries\large,  
    boxrule=0.5mm,              
    drop shadow,                
    sharp corners=downhill,   
    arc=3mm                 
]

The clause titled ``Automatically Collected Information'' explicitly states that the Application may collect certain information automatically, including ``Wi-Fi mac address in your area.'' This language confirms that the collection of the user’s MAC address is a permitted and described activity under the policy. The question asks whether the policy describes the collection of the user's MAC address by ad networks, analytics services, or other third parties. While the clause does not name specific third parties such as ad networks or analytics services, it does describe the collection of the MAC address as part of the Application’s automatic data collection. The phrase ``information about the way you use the Application'' further supports that data collection is ongoing and includes technical identifiers like the MAC address. The policy does not restrict or prohibit such collection, nor does it limit it to first-party use only. Furthermore, the clause refers to ``Automatic Data Collection and Advertising'' as a reference point for examples, implying that such collection includes third-party data sharing. Given that the MAC address is listed as one of the types of data collected, and that automatic collection often involves third-party services for analytics or advertising, the inclusion of ``Wi-Fi mac address'' in the list of collected data logically extends to its use by ad networks or analytics services, which are commonly engaged in such collection. Therefore, the policy does describe the collection of the user’s MAC address by third parties such as ad networks and analytics services, as the collection is both permitted and implicitly supported by the broader context of automatic data collection and advertising practices. Final Answer: Yes

\end{tcolorbox}

\section{Datasets}
\label{app:datasets}

\subsection{Public Benchmarks}
In the experiments, we use four publicly available datasets from LegalBench \cite{guha2023legalbench}:
\begin{itemize}
    \item \textbf{Consumer QA} is a benchmark that assesses whether models can interpret consumer contracts by answering yes/no questions about the legal implications of terms of service from popular online platforms. The dataset contains 400 annotated question–answer pairs.
    \item \textbf{Contracts QA} is a dataset consisting of contract clauses and a yes/no question, typically regarding whether the clause belongs to a certain provision category. The dataset contains 88 samples.
    \item \textbf{Sara Entailment} dataset evaluates a model’s ability to perform statutory reasoning by determining whether a legal rule applies to a fact pattern, using simplified tax‑law sections paired with an entailment question that require a binary yes/no answer. The dataset contains 276 examples.
    \item \textbf{Privacy Policy Entailment} dataset evaluates whether a model can identify privacy practices by determining if a given practice description entails a policy clause, formulated as a binary classification task. The dataset contains 4,343 examples.
\end{itemize}
All benchmarks are formatted as query–answer pairs with binary verdicts (yes/no or equivalent) to match our experimental setting.

\subsection{Real-World Datasets}
We include two proprietary real‑world datasets collected from authentic business scenarios within the financial industry, to assess practical utility under realistic document lengths and compliance‑oriented legal document review settings.

\begin{itemize}
    \item \textbf{Real-World POA} dataset is a document QA benchmark for reviewing power of attorney (POA) documents, where legal experts answer compliance questions with yes/no verdicts grounded in the underlying text. The dataset uses a fixed set of 13 questions and contains 780 examples in total. Example questions include "Does the document allow the agent to perform banking transactions?", "Does the document allow the agent to trade securities?", etc. Because POA documents tend to be long, we retrieve the top‑5 most similar snippets as the document context.
    \item \textbf{Real-World Trust} dataset is a document QA benchmark for reviewing trust-related documents, consisting of 12 fixed compliance questions with yes/no labels. Example questions include "Does the Trust authorize a trustee to charge fees?", "Is there more than one individuals named as trustee at the time of this trust creation?", etc. This dataset includes 468 examples. Similar to the POA setting, we retrieve the top‑5 most relevant snippets as context due to the length of the source documents.
\end{itemize}

\section{Implementation Details}

Our experimental framework is built upon the Hugging Face \texttt{transformers}, \texttt{TRL}, and \texttt{vLLM} libraries, with all models trained on NVIDIA A6000 GPUs. Due to the manageable scale of our 0.6B and 1.7B models, we performed full-parameter fine-tuning for both SFT and DPO stages. To address the risks of overfitting and model collapse associated with full-parameter updates on smaller architectures, we applied rigorous regularization. Specifically for DPO, we selected weight decay values in the range of $[1\times 10^{-5}, 1\times 10^{-3}]$ and learning rates between $[1\times 10^{-6}, 1\times 10^{-4}]$. Furthermore, we restricted DPO training to 1--3 epochs to maintain training stability and prevent reward hacking.

\newpage
\section{Prompts}
\label{app:prompts}
\subsection{Audit Agent Prompt}
\begin{tcolorbox}[
    enhanced,
    title={Agent System Prompt (Core Logic)}, 
    colframe=mypromptgrayheader,
    colback=mypromptbluebody,
    coltitle=white,
    fonttitle=\bfseries\large,
    boxrule=0.5mm,
    drop shadow,
    sharp corners=downhill,
    arc=3mm
]

\textbf{Role:} \\
You are a rigorous AI teaching assistant specializing in legal contract analysis.

\vspace{0.2cm}
\textbf{Core Objectives:}
\begin{enumerate}[leftmargin=*, nosep]
    \item \textbf{Diagnose:} Evaluate the correctness and reasoning of the student's answer against the ground truth.
    \item \textbf{Instruction Generation:} If an error exists, provide a specific, abstract instruction on how to \textit{reproduce} this logic error in a completely different legal context.
\end{enumerate}

\vspace{0.2cm}
\textbf{Internal Evaluation Process:}
\begin{enumerate}[leftmargin=*, nosep]
    \item Verify if the student's final answer matches the ground truth.
    \item Check reasoning for logical soundness (e.g., missed conditions, hallucinations).
    \item Classify any flaws using the provided \textit{Error Taxonomy}.
    \item Draft a \texttt{reproduction\_instruction} for the Teacher AI.
\end{enumerate}

\vspace{0.2cm}
\textbf{Reproduction Instruction Guidelines:} \\
When writing the instruction, you must be \textbf{Context-Agnostic} and \textbf{Actionable}. Do NOT mention specific entities or clauses.
\begin{itemize}[leftmargin=*, nosep]
    \item \textit{Example (Good):} "Identify a condition in the text that limits a right, and generate a response that treats the right as absolute by deliberately ignoring that condition."
    \item \textit{Example (Bad):} "Ignore the 5-day notice period in Clause 4."
\end{itemize}

\vspace{0.2cm}
\textbf{Output Format:} \\
Respond with a strict JSON object containing the evaluation status, error types, generic summary, and the reproduction instruction.
\end{tcolorbox}

The user prompt provides the agent with the specific context required for each evaluation instance. It is structured as a template that sequentially inputs the Contract text, the associated Question, and the Ground Truth (correct answer) to establish the evaluation standard. These are followed by the Student Answer that needs to be assessed. The prompt concludes with a directive triggering the agent to generate the output in the strict JSON format defined in the system prompt.

\subsection{Teacher Model Prompt}
\begin{tcolorbox}[
    enhanced,
    title={Teacher Model System Prompt for Generating Rejected Sample},
    colframe=mypromptgrayheader,
    colback=mypromptbluebody,
    coltitle=white,
    fonttitle=\bfseries\large,
    boxrule=0.5mm,
    drop shadow,
    sharp corners=downhill,
    arc=3mm
]

\textbf{Role \& Objective:} \\
You are an AI tutor acting as a student to create a flawed educational example. Your task is to generate an \textbf{incorrect} student answer.

\vspace{0.2cm}
\textbf{Core Mechanism:}
\begin{enumerate}[leftmargin=*, nosep]
    \item \textbf{Input Processing:} You will receive the \textit{Correct Answer} and specific \textit{Error Summaries}.
    \item \textbf{Flaw Embodiment:} You must generate a step-by-step reasoning process that naturally embodies the specified error (e.g., ignoring a condition) without explicitly stating "I am making an error."
    \item \textbf{Target Outcome:} Your reasoning must plausibly lead to the \textbf{opposite} of the Ground Truth.
\end{enumerate}

\vspace{0.2cm}
\textbf{Constraints:} \\
Do NOT mention the error summary or the ground truth in your output. Act entirely within the persona of the confused student.

\end{tcolorbox}

The user prompt constructs the simulation context by providing the Contract, the Question, and the Correct Answer (Ground Truth). Crucially, it injects the specific error parameters derived from the evaluation phase: the target Error Types, a Generic Error Description, and the specific Reproduction Instruction. These inputs serve as the blueprint for the model to construct the flawed reasoning path.

\begin{tcolorbox}[
    enhanced,
    title={Teacher Model User Prompt for Generating Rejected Sample (Core Constraint)},
    colframe=mypromptgrayheader,
    colback=mypromptbluebody,
    coltitle=white,
    fonttitle=\bfseries\large,
    boxrule=0.5mm,
    drop shadow,
    sharp corners=downhill,
    arc=3mm
]

\textbf{Generation Rules:}
\begin{enumerate}[leftmargin=*, nosep]
    \item \textbf{Immediate Reasoning:} Start the response immediately with the flawed step-by-step reasoning. Do NOT include any preamble or repetition of instructions.
    \item \textbf{Twist the Logic:} Plausibly embody the flaws listed in \texttt{Error Types}. Follow the \texttt{Reproduction Instruction} to manipulate the logic (e.g., if instructed to ignore a condition, simply fail to mention it).
    \item \textbf{Opposite Conclusion:} The reasoning must naturally lead to a final answer that is the \textbf{opposite} of the Correct Answer.
    \item \textbf{Formatting:} Strict adherence to the provided output structure is required.
    \item \textbf{Final Answer:} Conclude strictly with "Final Answer: Yes" or "Final Answer: No".
\end{enumerate}

\end{tcolorbox}

When generating the chosen sample, the prompt is similar except highlighting that the model needs to generate correct reaosning with an addtional input, which the rejected sample that the teacher model just generated. 

\subsection{Student Model Prompt}
\begin{tcolorbox}[
    enhanced,
    title={Student Model Prompt},
    colframe=mypromptgrayheader,
    colback=mypromptbluebody,
    coltitle=white,
    fonttitle=\bfseries\large,
    boxrule=0.5mm,
    drop shadow,
    sharp corners=downhill,
    arc=3mm
]

\textbf{System Role:} \\
You are an AI assistant specializing in legal contract analysis. Analyze contracts carefully, reference specific clauses, and provide step-by-step reasoning.

\vspace{0.2cm}
\textbf{Input Context:}
\begin{itemize}[leftmargin=*, nosep]
    \item \textbf{Contract:} \texttt{\{contract\}}
    \item \textbf{Question:} \texttt{\{question\}}
\end{itemize}

\vspace{0.2cm}
\textbf{Task Instructions:}
\begin{enumerate}[leftmargin=*, nosep]
    \item \textbf{Reference:} Explicitly identify and cite each clause or term relevant to the question.
    \item \textbf{Chain-of-Thought:} Provide a detailed, step-by-step explanation. For each step:
    \begin{itemize}[nosep]
        \item Quote/Summarize the specific contract part (e.g., "Clause 2.1 states...").
        \item Explain its logical impact on the answer.
    \end{itemize}
    \item \textbf{Final Answer:} Conclude clearly on a new line with "Final Answer: Yes" or "Final Answer: No".
\end{enumerate}

\end{tcolorbox}

\subsection{Judge Model Prompt}
\begin{tcolorbox}[
    enhanced,                   
    title={Judge System Prompt},
    colframe=mypromptgrayheader,         
    colback=mypromptbluebody,             
    coltitle=white,             
    fonttitle=\bfseries\large,  
    boxrule=0.5mm,              
    drop shadow,                
    sharp corners=downhill,   
    arc=3mm                 
]
You are a strict legal reasoning judge. Field definitions: question = the claim/question to evaluate; contract = the governing legal text/context only; ground\_truth = the gold final answer for the question under the contract. Given question, contract, ground\_truth, and a model response, decide whether the model response contains ANY legal error (factual, logical, interpretive, or conclusion mismatch). Hard rule: if the model's final answer does not match ground\_truth, the judgment must be incorrect. Output exactly one word: correct or incorrect.
\end{tcolorbox}

\vspace{1em}

\begin{tcolorbox}[
    enhanced,                   
    title={Judge User Prompt Template},
    colframe=mypromptgrayheader,         
    colback=mypromptbluebody,             
    coltitle=white,             
    fonttitle=\bfseries\large,  
    boxrule=0.5mm,              
    drop shadow,                
    sharp corners=downhill,   
    arc=3mm                 
]
Question:\\
\{question\}

Legal Context:\\
\{contract\}

Ground Truth Answer:\\
\{ground\_truth\}

Model Response To Judge:\\
\{model\_generation\}

Instruction:\\
- Meaning reminder:\\
  question = what needs to be judged; contract = legal basis; ground\_truth = gold final answer\\
- First extract the model's final answer from the model response\\
- If the extracted final answer does NOT match ground\_truth, output: incorrect (mandatory)\\
- If there is any legal error anywhere in the model response, output: incorrect\\
- Otherwise output: correct\\
- Output one word only.
\end{tcolorbox}


\onecolumn
\appendix

\end{document}